\documentclass{article}

\usepackage{colt11e}

\usepackage{amssymb}
\usepackage{amsmath}

\usepackage{graphicx}
\usepackage{subfigure}

\usepackage[round,comma]{natbib}
\renewenvironment{proof}[1][Proof]{\noindent{\bf #1:}}{\qed\medskip}

\title{PAC-Bayesian Analysis of Martingales\\and Multiarmed Bandits}

\author{Yevgeny Seldin\\
Max Planck Institute\\
T\"{u}bingen, Germany\\
\texttt{\small seldin@tuebingen.mpg.de}
\And 
Fran\c{c}ois Laviolette\\
Universit\'{e} Laval \\
Qu\'{e}bec, Canada \\
\texttt{\small francois.laviolette@ift.ulaval.ca}
\And
John Shawe-Taylor\\
University College London\\
\texttt{\small jst@cs.ucl.ac.uk}
\AND
Jan Peters\\
Max Planck Institute\\
T\"{u}bingen, Germany\\
\texttt{\small jan.peters@tuebingen.mpg.de}
\And
Peter Auer\\
Chair for Information Technology\\
University of Leoben, Austria\\
\texttt{\small auer@unileoben.ac.at}
}

\renewcommand{\cite}{\citep}

\begin{document}

\maketitle

\begin{abstract}
We present two alternative ways to apply PAC-Bayesian analysis to sequences of dependent random variables. The first is based on a new lemma that enables to bound expectations of convex functions of certain dependent random variables by expectations of the same functions of independent Bernoulli random variables. This lemma provides an alternative tool to Hoeffding-Azuma inequality to bound concentration of martingale values. Our second approach is based on integration of Hoeffding-Azuma inequality with PAC-Bayesian analysis. We also introduce a way to apply PAC-Bayesian analysis in situation of limited feedback. We combine the new tools to derive PAC-Bayesian generalization and regret bounds for the multiarmed bandit problem. Although our regret bound is not yet as tight as state-of-the-art regret bounds based on other well-established techniques, our results significantly expand the range of potential applications of PAC-Bayesian analysis and introduce a new analysis tool to reinforcement learning and many other fields, where martingales and limited feedback are encountered.
\end{abstract}

\section{Introduction}

PAC-Bayesian analysis was introduced over a decade ago \cite{STW97,ST+98,McA98,See02} and has since made a significant contribution to the analysis and development of supervised learning methods. The power of PAC-Bayesian approach lies in the successful marriage of flexibility and intuitiveness of Bayesian models with the rigor of PAC analysis. PAC-Bayesian bounds provide an explicit and often intuitive and easy-to-optimize trade-off between model complexity and empirical data fit, where the complexity can be nailed down to the resolution of individual hypotheses via the prior definition. The PAC-Bayesian analysis was applied to derive generalization bounds and new algorithms for linear classifiers and maximum margin methods \cite{LST02, McA03b, GLLM09}, structured prediction \cite{McA07}, and clustering-based classification models \cite{ST10}, to name just a few. However, the application of PAC-Bayesian analysis beyond the supervised learning domain remained surprisingly limited. In fact, the only additional domain known to us is density estimation \cite{ST10, HST10}. 

Even within supervised learning the applications of PAC-Bayesian analysis were restricted to i.i.d. data for a long time. The issue of treating non-independent samples was partially addressed only recently by \citet{RSS10} and \citet{LLST10} (their approaches are also suitable for density estimation \cite{HST10}. The solution of \citet{RSS10} essentially boils down to breaking the sample into independent (or almost independent) subsets (which also reduces the effective sample size to the number of independent subsets). Such an approach is inapplicable to martingales due to strong dependence of the cumulative sum on all of its components. \citet{LLST10} employed Hoeffding's canonical decomposition of U-statistics into forward martingales and applied PAC-Bayesian analysis directly to these martingales. Our second approach to handling sequences of dependent samples by combining PAC-Bayesian analysis with Hoeffding-Azuma inequality is based on similar ideas. Our first approach to sequences of dependent samples is based on the new lemma that allows to bound expectations of functions of certain sequentially dependent random variables by expectations of the same functions of independent random variables.

One of the most prominent and important fields of application of martingales is reinforcement learning. Some potential advantages of applying PAC-Bayesian analysis in reinforcement learning were recently pointed out by several researchers, including \citet{TP10} and \citet{FP10}. \citet{TP10} suggested that the mutual information between states and actions in a policy can be used as a natural regularizer in reinforcement learning. They showed that regularization by mutual information can be incorporated into Bellman equations and therefore can be computed efficiently. Tishby and Polani conjectured that PAC-Bayesian analysis can be applied to justify such form of regularization and provide generalization guarantees for it.

\citet{FP10} suggested a PAC-Bayesian analysis of batch reinforcement learning. They used the analysis to design an algorithm that is able to leverage the prior knowledge when it is informative and confirms the data distribution and ignores it when it is irrelevant. In the first case Bayesian learning algorithms perform well and in the second case PAC learning algorithms perform better, whereas Fard and Pineau showed that their algorithm performs on par with the best out of the two in all situations. However, the analysis of Fard and Pineau does not address the exploration-exploitation trade-off, which is the key feature of reinforcement learning. In their batch analysis they assume that every action was sampled in every state some minimal number of times and the bound decreases at the rate of a square root of the minimum over states and actions of the number of times an action was sampled in a state. Clearly, such an analysis is not applicable in online setting, since we do not want to sample ``bad'' actions many times, but then the bound does not improve with time.

One of the reasons for the difficulty of applying PAC-Bayesian analysis to address the exploration-exploitation trade-off is the limited feedback (the fact that we only observe the reward for the action taken, but not for all the rest). In supervised learning (and also in density estimation) the empirical error for each hypothesis within a hypotheses class can be evaluated on all the samples and therefore the size of the sample available for evaluation of all the hypotheses is the same (and usually relatively large). In the situation of limited feedback the sample from one action cannot be used to evaluate another action (that is the reason why the bound of \citet{FP10} depends on the minimum of the number of times any action was taken in any state, which is the minimal sample size available for evaluation of all state-action pairs). In online setting the sample size of ``bad'' actions has to increase sublinearly in the number of game rounds, which results in slow or even no convergence of the bound. We resolve this issue by applying weighted sampling strategy \cite{SB98}, which is commonly used in the analysis of non-stochastic bandits \cite{ACB+02}, but has not been applied to the analysis of stochastic bandits previously.

The usage of weighted sampling introduces two new difficulties. One is the dependence between the samples: the rewards we observe influence the distribution over actions we play and through this distribution influence the variance of the subsequent weighted sample variables. We handle this dependence using our new PAC-Bayesian approaches to sequences of dependent variables. At the moment both approaches yield comparable bounds, however each of the approaches has its own potential advantages that can be exploited in future work.

The second problem introduced by weighted sampling is the growing variance of the weighted sample variables. Martingale bounding techniques used in this work do not enable to take full control over the variance, which explains the gap between our results and state-of-the-art bounds for mutliarmed bandits \cite{ACBF02, AO10}. Tighter bounds can be achieved by combining PAC-Bayesian analysis with Bernstein-type inequality for martingales \cite{BLL+10}. Such a combination will be presented in future work.

The subsequent sections are organized as follows: Section \ref{sec:main} surveys the main results of the paper, Section \ref{sec:MartinLemma} presents our bound on expectation of convex functions of sequentially dependent random variables and illustrates its application to derivation of an alternative to Hoeffding-Azuma inequality, Section \ref{sec:PAC-B} provides a PAC-Bayesian analysis of the weighted sampling strategy based on the bound from Section \ref{sec:MartinLemma}, Section \ref{sec:Martin} provides PAC-Bayesian analysis of the weighted sampling strategy based on martingales, Section \ref{sec:regret} derives a regret bound for the multiarmed bandit problem, and Section \ref{sec:disc} concludes the results.

\section{Main Results}
\label{sec:main}

One of the foundation stones of our paper is the following lemma that enables to bound expectations of convex functions of certain sequentially dependent random variables by expectations of the same functions of independent Bernoulli random variables. The lemma generalizes a preceding result of \citet{Mau04} for independent random variables and might have a wide interest on its own right far beyond the PAC-Bayesian analysis. The lemma can be used to derive an alternative to Hoeffding-Azuma inequality \cite{Hoe63, Azu67}. This alternative can be much tighter in certain situations (see our derivation and discussion of Lemma \ref{lem:Azu2} in the next section).

\begin{lemma}
\label{lem:Martin}
Let $X_1,..,X_N$ be dependent random variables belonging to the $[0,1]$ interval and distributed by $p(x_i|X_1,..,X_{i-1})$, such that $\mathbb E [X_i|X_1,..,X_{i-1}] = p$ for all $i$. Let $Y_1,..,Y_N$ be independent Bernoulli random variables, such that $\mathbb E Y_i = p$ for all $i$. Then for any convex function $f:[0,1]^N \rightarrow \mathbb R:$
\[
\mathbb E f(X_1,..,X_N) \leq \mathbb E f(Y_1,..,Y_N).
\]
\end{lemma}

We present the subsequent results in the context of the multiarmed bandit problem, which is probably the most common problem in machine learning, where sequentially dependent variables are encountered. Let ${\cal A}$ be a set of actions (arms) of size $|{\cal A}| = K$ and let $a \in {\cal A}$ denote the actions. Denote by $R(a)$ the expected reward of action $a$. Let $\pi_t$ be a distribution over ${\cal A}$ that is played at round $t$ of the game. Let $\{A_1,A_2,...\}$ be the sequence of actions played independently at random according to $\{\pi_1,\pi_2,...\}$ respectively. Let $\{R_1,R_2,...\}$ be the sequence of observed rewards. Denote by ${\cal T}_t = \left \{\{A_1,..,A_t\},\{R_1,..,R_t\}\right\}$ the set of taken actions and observed rewards up to round $t$ (by definition ${\cal T}_{t-1} \subset {\cal T}_t$).

For $t\geq 1$ and $a \in \{1,..,K\}$ define a set of indicator random variables $\{I_t^a\}_{t,a}$:
\[
I_t^a = \left \{ \begin{array}{ll}1,&\mbox{if}~A_t=a\\0,&\mbox{otherwise.}\end{array} \right .
\]
Define a set of random variables $R_t^a = \frac{1}{\pi_t(a)}I_t^a R_t$. In other words:
\[
R_t^a = \left \{ \begin{array}{cl}\frac{1}{\pi_t(a)}R_t,&\mbox{if}~A_t=a\\0,&\mbox{otherwise.}\end{array} \right .
\]

Define: $\hat R_t(a) = \frac{1}{t} \sum_{\tau=1}^t R_\tau^a$. For a distribution $\rho$ over ${\cal A}$ define $R(\rho) = \mathbb E_{\rho(a)} R(a)$ and $\hat R_t(\rho) = \mathbb E_{\rho(a)} \hat R_t(a)$. 

For two distributions $\rho$ and $\mu$, let $KL(\rho\|\mu)$ denote the KL-divergence between $\rho$ and $\mu$. For two Bernoulli random variables with biases $p$ and $q$ let $kl(p\|q) = p \ln \frac{p}{q} + (1-p) \ln \frac{1-p}{1-q}$ be an abbreviation for $KL([p, 1-p]\|[q, 1-q])$.

We present two alternative results, the first applies Lemma \ref{lem:Martin} to handle sequences of dependent random variables and the second is based on combination of PAC-Bayesian analysis with Hoeffding-Azuma inequality. Then we compare the results and present a regret bound for the multiarmed bandit problem based on the first solution.

\subsection{PAC-Bayesian Analysis of Sequentially Dependent Variables Based on Lemma \ref{lem:Martin}}

Our first PAC-Bayesian theorem provides a bound on the divergence between $\hat R_t(\rho_t)$ and $R(\rho_t)$ for any playing strategy $\rho_t$ throughout the game.

\begin{theorem}
\label{thm:PAC-Bayes-Bandits}
For any sequence of sampling distributions $\{\pi_1,\pi_2,...\}$ that are not zero for any $a \in {\cal A}$, where $\pi_t$ can depend on ${\cal T}_{t-1}$, and for any sequence of ``reference'' $($``prior''$)$ distributions $\{\mu_1,\mu_2,...\}$ over ${\cal A}$, such that $\mu_t$ is independent of ${\cal T}_t$ $($but can depend on $t$$)$, for all possible distributions $\rho_t$ given $t$ and for all $t \geq 1$ simultaneously with probability greater than $1-\delta$$:$
\begin{equation}
kl(\pi_t^{lmin} \hat R_t(\rho_t)\|\pi_t^{lmin} R(\rho_t)) \leq \frac{KL(\rho_t\|\mu_t) + 3 \ln (t+1) - \ln \delta}{t},
\label{eq:PAC-Bayes-Bandits}
\end{equation}
where
\[
\pi_t^{lmin} \leq \min_{\substack{a,\\1 \leq \tau \leq t}} \pi_\tau(a).
\]
\end{theorem}

The number $\pi_t^{lmin}$ lower bounds sampling probabilities for all the actions up to time $t$ ($lmin$ stands for ``left minimum'' or minimum of $\pi_\tau(a)$ up to [``left to''] time $t$).

The KL-divergence $kl(p\|q)$ bounds the absolute difference between $p$ and $q$ as
\begin{equation}
\label{eq:kl-L1}
|p-q| \leq \sqrt{kl(p\|q)/2}
\end{equation}
\cite{CT91}. Combined with \eqref{eq:PAC-Bayes-Bandits} this relation yields (with probability greater than $1-\delta$):
\begin{equation}
\left |R(\rho_t) - \hat R_t(\rho_t) \right | \leq \frac{1}{\pi_t^{lmin}} \sqrt{\frac{KL(\rho_t\|\mu_t) + 3 \ln (t+1) - \ln \delta}{2t}}.
\label{eq:PAC-Bayes-Bandits-L1}
\end{equation}

\subsection{Combination of PAC-Bayesian Analysis with Hoeffding-Azuma Inequality}

The result presented next is based on a combination of PAC-Bayesian analysis with Hoeffding-Azuma inequality. We introduce one more definition:
\[
\hat R_t^{w^t}(a) = \frac{\sum_{\tau = 1}^t w^t_\tau R_\tau^a}{\sum_{\tau=1}^t w^t_{\tau}},
\]
where $w^t_\tau \geq 0$ for all $t$ and $\tau$ and $\sum_{\tau=1}^t w^t_{\tau} > 0$ for all $t$. $\hat R_t^{w^t}(a)$ is a weighted sum of the samples. For a special case, where $w^t_\tau = \frac{1}{t}$ for all $\tau$, $\hat R_t^{w^t}(a) = \hat R_t(a)$.

\begin{theorem}
\label{thm:PAC-Bayes-Martin}
For any sequence of sampling distributions $\{\pi_1,\pi_2,...\}$ that are not zero for any $a \in {\cal A}$, where $\pi_t$ can depend on ${\cal T}_{t-1}$, and for any sequence of ``reference'' $($``prior''$)$ distributions $\{\mu_1,\mu_2,...\}$ over ${\cal A}$, such that $\mu_t$ is independent of ${\cal T}_t$ $($but can depend on $t$$)$, for any sequence of positive parameters $\{\lambda_1,\lambda_2,...\}$ and for any sequence of weighting vectors $\{w^1,w^2,...\}$, such that $\lambda_t$ and $w^t$ are independent of ${\cal T}_t$ $($but can depend on $t$$)$, for all possible distributions $\rho_t$ given $t$ and for all $t \geq 1$ simultaneously with probability greater than $1-\delta$$:$
\begin{equation}
\left | \hat R_t^{w^t}(a) - R(a) \right | \leq \frac{KL(\rho_t\|\mu_t) + \frac{1}{2}\lambda_t^2 \sum_{\tau = 1}^t \left (\frac{w^t_\tau}{\pi_\tau^{min}} \right )^2 + 2 \ln(t+1) + \ln \frac{2}{\delta}}{\lambda_t \sum_{\tau = 1}^t w^t_\tau},
\label{eq:PAC-Bayes-Martin}
\end{equation}
where
\[
\pi_t^{min} \leq \min_a \pi_t(a).
\]
\end{theorem}

For the special case $w^t_\tau = \frac{1}{t}$ we obtain that with probability greater than $1 - \delta$:
\begin{equation}
\left | \hat R_t(a) - R(a) \right | \leq \frac{KL(\rho_t\|\mu_t) + \frac{1}{2}\frac{\lambda_t^2}{t^2} \sum_{\tau = 1}^t \frac{1}{\left (\pi_\tau^{min} \right )^2} + 2 \ln(t+1) + \ln \frac{2}{\delta}}{\lambda_t}.
\label{eq:PAC-Bayes-Martin-hatR}
\end{equation}
By taking
\[
\lambda_t = \sqrt{2 t^2 \left (2 \ln (t+1) + \ln \frac{2}{\delta} \right ) / \left ( \sum_{\tau = 1}^t \frac{1}{\left (\pi_\tau^{min} \right )^2} \right )}
\]
we obtain:
\begin{equation}
\left | \hat R_t(a) - R(a) \right | \leq \sqrt{\frac{\frac{1}{t} \left ( \sum_{\tau = 1}^t \frac{1}{\left (\pi_\tau^{min} \right )^2} \right )}{2t}} \left (\frac{KL(\rho_t\|\mu_t)}{\sqrt{\ln (t+1) + \ln \frac{2}{\delta}}} + \sqrt{\ln (t+1) + \ln \frac{2}{\delta}} \right ).
\label{eq:compare}
\end{equation}

\subsection{Comparison of Theorem \ref{thm:PAC-Bayes-Bandits} with Theorem \ref{thm:PAC-Bayes-Martin}}
It is interesting to compare Theorems \ref{thm:PAC-Bayes-Bandits} and \ref{thm:PAC-Bayes-Martin} resulting from the two different approaches. Inequality \eqref{eq:PAC-Bayes-Bandits-L1} depends on $\frac{1}{\pi_t^{lmin}}= \max_{1 \leq \tau \leq t} \left \{\frac{1}{\pi_\tau^{min}} \right \}$, whereas \eqref{eq:compare} depends on $\sqrt{\frac{1}{t} \sum_{\tau=1}^t \frac{1}{\left (\pi_\tau^{min} \right )^2}}$. If $\pi_\tau^{min}$ are approximately equal for all $\tau$, then the two terms are approximately identical. However, a single small value of $\pi_\tau^{min}$ can increase the value of $\frac{1}{\pi_t^{lmin}}$ significantly for all $t \geq \tau$, while its relative contribution to the average of $\frac{1}{\left (\pi_\tau^{min} \right )^2}$ will decrease with time. This property provides an advantage to Theorem \ref{thm:PAC-Bayes-Martin}. On the other hand, the stronger $kl$ form \eqref{eq:PAC-Bayes-Bandits} of Theorem \ref{thm:PAC-Bayes-Bandits} can potentially be an advantage for the bound based on Lemma \ref{lem:Martin}, but we did not exploit it in this work.

Since for our choice of sampling strategy $\frac{1}{\pi_t^{lmin}} \approx \sqrt{\frac{1}{t} \sum_{\tau=1}^t \frac{1}{\left (\pi_\tau^{min} \right )^2}}$ up to small constants, we present a regret bound based on Theorem \ref{thm:PAC-Bayes-Bandits} only. A regret bound based on Theorem \ref{thm:PAC-Bayes-Martin} can be derived in a similar way and is identical to the bound presented below up to small constants.

\subsection{Regret Bound for Multiarmed Bandits}

We applied Theorem \ref{thm:PAC-Bayes-Bandits} to derive the following regret bound for the multiarmed bandit problem.

\begin{theorem}
\label{thm:regret}
For $t < K^3$ let $\pi_t(a) = \frac{1}{K}$ for all $a$. Let $\gamma_t = K^{1/4}t^{1/4}$ and $\varepsilon_t = K^{-1/4}t^{-1/4}$ and for $t \geq (K^3 - 1)$ let
\begin{equation}
\pi_{t+1}(a) = \tilde \rho_t^{_{exp}}(a) = (1 - K \varepsilon_{t+1}) \rho_t^{_{exp}}(a) + \varepsilon_{t+1},
\label{eq:tilderho}
\end{equation}
where
\begin{equation}
\rho_t^{_{exp}}(a) = \frac{1}{Z(\rho_t^{_{exp}})} e^{\gamma_t \hat R_t(a)}
\label{eq:rho}
\end{equation}
and
\[
Z(\rho_t^{_{exp}}) = \sum_a e^{\gamma_t \hat R_t(a)}.
\]
Then for $t \geq K^3$ the per-round regret $R(a^*) - R(\tilde \rho_t^{_{exp}})$ $($where $a^*$ is the best action$)$ is bounded by$:$
\[
R(a^*) - R(\tilde \rho_t^{_{exp}}) \leq \frac{K^{3/4}}{(t+1)^{1/4}}
\left (
2.5 + \sqrt{\frac{\ln(K) + 3 \ln(t+1) - \ln \delta}{2K}}+ \sqrt{\frac{3 \ln(t+1) - \ln \delta}{2K}} 
\right )
\]
with probability greater than $1-\delta$ for all rounds $t$ simultaneously. This translates into a total regret of $\tilde O(K^{3/4}t^{3/4})$ $($where $\tilde O$ hides logarithmic factors$)$.
\end{theorem}

Note that $\varepsilon_t$ bounds $\pi_t(a)$ from below for all $a$ and $t \geq K^3$. Furthermore, since $\varepsilon_t$ is a decreasing sequence it actually bounds $\pi_\tau(a)$ from below for all $a$ and $\tau \leq t$. Hence, for the prediction strategy selected in Theorem \ref{thm:regret} and for $t \geq K^3$ we can substitute $\pi_t^{lmin}$ with $\varepsilon_t$ in \eqref{eq:PAC-Bayes-Bandits} and \eqref{eq:PAC-Bayes-Bandits-L1}.

\section{Proof of Lemma \ref{lem:Martin} and an Example of its Application}
\label{sec:MartinLemma}

We start with the proof of Lemma \ref{lem:Martin} and then illustrate how it can be applied to martingales.

\begin{proof}[Proof of Lemma \ref{lem:Martin}]
The proof follows the lines of the proof of Lemma 3 in \citet{Mau04}. Any point $\bar x = (x_1,..,x_N) \in [0,1]^N$ can be written as a convex combination of the extreme points $\bar \eta = (\eta_1,..,\eta_N) \in \{0,1\}^N$ in the following way:
\[
\bar x = \sum_{\bar \eta \in \{0,1\}^N} \left (\prod_{i:\eta_i = 0} (1 - x_i) \prod_{i:\eta_i = 1} x_i \right ) \bar \eta.
\]
Convexity of $f$ therefore implies
\begin{equation}
f(\bar x) \leq \sum_{\bar \eta \in \{0,1\}^N} \left (\prod_{i:\eta_i = 0} (1 - x_i) \prod_{i:\eta_i = 1} x_i \right ) f(\bar \eta),
\label{eq:convexity}
\end{equation}
with equality if $\bar x \in \{0,1\}^N$. At the next step \citet{Mau04} uses independence of $X_i$-s, whereas we use the fact that their conditional expectation is constant. Taking expectation of both sides of \eqref{eq:convexity} we obtain:
\begin{align}
\mathbb E&_{X_1,..,X_N} [f(\bar X)] \leq \mathbb E_{X_1,..,X_N} \left [ \sum_{\bar \eta \in \{0,1\}^N} \left (\prod_{i:\eta_i = 0} (1 - X_i) \prod_{i:\eta_i = 1} X_i \right ) f(\bar \eta) \right ]\notag\\
&= \sum_{\bar \eta \in \{0,1\}^N} \mathbb E_{X_1,..,X_N} \left [ \left (\prod_{i:\eta_i = 0} (1 - X_i) \prod_{i:\eta_i = 1} X_i \right ) \right ] f(\bar \eta) \notag\\
&= \sum_{\bar \eta \in \{0,1\}^N} \mathbb E_{X_1,..,X_{N-1}} \left [ \mathbb E_{X_N} \left [ \left . \left (\prod_{i:\eta_i = 0} (1 - X_i) \prod_{i:\eta_i = 1} X_i \right ) \right | X_1,..,X_{N-1}\right ]\right ]f(\bar \eta)\notag\\
&= \sum_{\bar \eta \in \{0,1\}^N} \mathbb E_{X_1,..,X_{N-1}} \left [ \begin{array}{l}\left (\prod_{i:\eta_i = 0, i<N} (1 - X_i) \prod_{i:\eta_i = 1, i<N} X_i \right ) \\ \quad \cdot \mathbb E_{X_N} \left [(1 -\eta_{_N} ) (1-X_{_N}) + \eta_{_N} X_{_N} | X_1,..,X_{N-1}\right ]\end{array} \right ]f(\bar \eta)\notag\\
&= \sum_{\bar \eta \in \{0,1\}^N} \mathbb E_{X_1,..,X_{N-1}} \left [ 
\left (\prod_{i:\eta_i = 0, i<N} (1 - X_i) \prod_{i:\eta_i = 1, i<N} X_i \right )
\cdot  \left [(1 - \eta_{_N}) (1-p) + \eta_{_N} p \right ]
\right ]f(\bar \eta)\notag\\
&= ... \label{eq:induction}\\
&= \sum_{\bar \eta \in \{0,1\}^N} \left (\prod_{i:\eta_i = 0} (1 - p) \prod_{i:\eta_i = 1} p \right ) f(\bar \eta)\notag\\
&= \mathbb E_{Y_1,..,Y_N} [f(\bar Y)].\notag
\end{align}
In \eqref{eq:induction} we apply induction in order to replace $X_i$-s by $p$, one-by-one from the last to the first, same way we did it for $X_N$.
\end{proof}

\subsection{Application to Martingales}

We apply Lemma \ref{lem:Martin} to derive an alternative to Hoeffding-Azuma inequality. The derivation is based on Markov's inequality and a concentration result for independent Bernoulli variables provided below.

\begin{lemma}[Markov's inequality]
\label{lem:Markov}
For a random variable $X \geq 0$ with probability greater than $1-\delta$$:$
\begin{equation}
X \leq \frac{1}{\delta} \mathbb E X.
\end{equation}
\end{lemma}

The concentration result for independent Bernoulli variables is based on the method of types in information theory \cite{CT91}. Its proof can be found in \citet{See03},\citet{Ban06}, or \citet{ST10}.\footnote{It is possible to prove even stronger result of a form $\sqrt N \leq \mathbb E_{X_1,..,X_N}e^{N kl(\hat S\|S)} \leq 2 \sqrt N$ for $N\geq8$ using Stirling's approximation of the factorial \cite{Mau04}. For simplicity we use \eqref{eq:Laplace}.}
\begin{lemma}
\label{lem:Laplace}
Let $X_1,..,X_N$ be i.i.d. Bernoulli random variables. Let $\hat S = \frac{1}{N}\sum_{i=1}^N X_i$ be their empirical average and $S = \mathbb E X_i$ the expected value. Then$:$
\begin{equation}
\mathbb E_{X_1,..,X_N} [e^{N kl(\hat S\|S)}] \leq N+1. 
\label{eq:Laplace}
\end{equation}
\end{lemma}

Since KL-divergence is a convex function \cite{CT91} and exponent is convex and non-decreasing, $e^{N kl(\hat S\|S)}$ is also a convex function. Therefore, by Lemma \ref{lem:Martin} we obtain that Lemma \ref{lem:Laplace} also holds for $X_1,..,X_N$ that belong to the $[0,1]$ interval and are sequentially dependent on each other as long as their conditional expectation $\mathbb E [X_i|X_1,..,X_{i-1}]$ is identical.

\subsubsection*{Alternative to Hoeffding-Azuma Inequality Based on Lemmas \ref{lem:Martin} and \ref{lem:Laplace}}

Now we are ready to present our alternative to Hoeffding-Azuma's inequality. 
\begin{lemma}
\label{lem:Azu2}
Let $X_1,..,X_N$ be a martingale difference sequence $($meaning that $\mathbb E [X_i|X_1,..,X_{i-1}] = 0$$)$, such that $X_i \in [a_i, b_i]$ for an arbitrary $a_i \leq 0$ and $b_i \geq 0$. Let $S_1,..,S_N$ be a martingale, where $S_j = \sum_{i=1}^j X_i$. Let $a = \min_i a_i$ and $b = \max_i b_i$ and let $Z_i = (X_i - a) / (b - a)$. Then with probability greater than $1-\delta$ the following holds simultaneously$:$
\begin{equation}
\label{eq:Azukl}
kl \left (\frac{1}{N} \sum_{i=1}^N Z_i \Arrowvert \frac{-a}{b-a} \right ) \leq \frac{\ln\frac{N+1}{\delta}}{N}
\end{equation}
and
\begin{equation}
\label{eq:AzuklL1}
|S_N| \leq (b - a) \sqrt{\frac{1}{2} N \ln \frac{N+1}{\delta}}.
\end{equation}
\end{lemma}

\begin{proof}[Proof of Lemma \ref{lem:Azu2}]
By definition of $Z_i$ we have $Z_i \in [0,1]$ and $\mathbb E [Z_i|Z_1,..,Z_{i-1}] = \frac{-a}{b-a}$ is identical for all $Z_i$. Hence, by Markov's inequality and combination of Lemma \ref{lem:Martin} with Lemma \ref{lem:Laplace} with probability greater than $1-\delta$:
\[
e^{N kl(\frac{1}{N} \sum_{i=1}^N Z_i \| \frac{-a}{b-a})} \leq \frac{1}{\delta} \mathbb E_{Z_1,..,Z_N} [e^{N kl(\frac{1}{N} \sum_{i=1}^N Z_i \| \frac{-a}{b-a})}] \leq \frac{N+1}{\delta}.
\]
Taking logarithm and normalizing by $N$ yields \eqref{eq:Azukl}.

By relation \eqref{eq:kl-L1} between $L_1$-norm and KL-divergence \eqref{eq:Azukl} yields:
\[
\left |\frac{1}{N} \sum_{i=1}^N Z_i - \frac{-a}{b-a} \right | \leq \sqrt{\frac{\ln\frac{N+1}{\delta}}{2N}}.
\]
From definitions, $X_i = (b-a) Z_i + a$ and $S_N = (b - a) \sum_{i=1}^N Z_i + N a$. Simple algebraic manipulations yield \eqref{eq:AzuklL1}.
\end{proof}

\subsubsection*{Comparison with Hoeffding-Azuma Inequality}

It is instructive to compare Lemma \ref{lem:Azu2} with Hoeffding-Azuma inequality, which we cite below for the comparison \cite{Azu67, CBL06}.
\begin{lemma}[Hoeffding-Azuma Inequality]
\label{lem:Azu}
Let $S_1,..,S_N$ be a zero-mean martingale satisfying $S_i - S_{i-1} \in [a_i,b_i]$, then for any $\lambda > 0$$:$
\[
\mathbb E [e^{\lambda S_N}] \leq e^{(\lambda^2 / 8) \sum_{i=1}^N (b_i - a_i)^2}. 
\]
\end{lemma}

It is easy to verify, using the same procedure we applied before, that Lemma \ref{lem:Azu} implies that with probability greater than $1-\delta$:
\[
|S_N| \leq \frac{\frac{1}{8}\lambda^2 \sum_{i=1}^N (b_i - a_i)^2 + \ln \frac{2}{\delta}}{\lambda}
\]
and that the above expression is minimized by $\lambda = \sqrt{8 \ln \frac{2}{\delta} / \sum_{i=1}^N (b_i - a_i)}$ yielding:
\begin{equation}
|S_N| \leq \sqrt{\frac{1}{2} \left (\sum_{i=1}^N (b_i - a_i)^2 \right ) \ln \frac{2}{\delta}}.
\label{eq:ba}
\end{equation}
In a special case, where $a_i = a$ for all $i$ and $b_i = b$ for all $i$, this further simplifies to:
\[
|S_N| \leq (b - a) \sqrt{\frac{1}{2} N \ln \frac{2}{\delta}}.
\]

Now we are ready to make the comparison. If $a_i$-s and $b_i$-s are equal (or almost equal) for all $i$, inequality \eqref{eq:AzuklL1} matches Hoeffding-Azuma inequality up to $\ln(N+1)$ factor (which can also be halved by using a tighter bound in \eqref{eq:Laplace}). If $a_i$-s and $b_i$-s are not identical, inequality \eqref{eq:AzuklL1} can be potentially much worse, since a single large $(b_i - a_i)$ term will permanently increase $(b-a)$, but its relative contribution to \eqref{eq:ba} will decrease with the increase of $N$. However, when the empirical average is close to lower or upper limit of the domain interval the $kl$ form of Lemma \ref{lem:Azu2} in equation \eqref{eq:Azukl} is much tighter than the relaxed $L_1$ norm form in equation \eqref{eq:AzuklL1} \cite{McA03b}. Therefore, in situations, where the analysis can be carried out using the $kl$ form of the bound, it might be preferable.

\section{Proof of Theorem \ref{thm:PAC-Bayes-Bandits} (PAC-Bayesian Bound Based on Lemma \ref{lem:Martin})}
\label{sec:PAC-B}

Our proof uses the following lemma, which lays at the basis of PAC-Bayesian analysis from its inception and takes its roots back in information theory and statistical physics \cite{DV75, DE97, Gra11, Ban06}. The lemma allows to relate all posterior distributions $\rho$ to a single prior distribution $\mu$.

\begin{lemma}
\label{lem:PAC-Bayes}
For any measurable function $\phi(h)$ on ${\cal H}$ and any distributions $\mu(h)$ and $\rho(h)$ on ${\cal H}$, we have$:$
\begin{equation}
\mathbb E_{\rho(h)}[\phi(h)] \leq KL(\rho\|\mu) + \ln \mathbb E_{\mu(h)}[ e^{\phi(h)}].
\label{eq:PAC-Bayes}
\end{equation}
\end{lemma}

\begin{proof}[Proof of Theorem \ref{thm:PAC-Bayes-Bandits}]
First, we show that $R(a) = \mathbb E_{{\cal T}_t}[\hat R_t(a)]$. Let $p(r|a)$ be the distribution of the reward for playing arm $a$ and let $R^a$ be a random variable distributed according to $p(r|a)$. Then for any $t$:
\begin{align}
R(a) &= \mathbb E_{p(r|a)} [R^a] = \mathbb E_{p(r|a)} \left [\pi_t(a) \frac{1}{\pi_t(a)} R^a \right ] = \mathbb E_{p(r|a)} \mathbb E_{\pi_t(a)} \left [\frac{1}{\pi_t(a)} I_t^a R^a \right ]\notag\\
 &= \mathbb E_{p(r|a),\pi_t(a)} \left [\frac{1}{\pi_t(a)} I_t^a R_t \right ] = \mathbb E_{p(r|a),\pi_t(a)} [R_t^a],\label{eq:1-1}
\end{align}
where \eqref{eq:1-1} holds since if $I_t^a=1$, then $R_t$ is distributed by $p(r|a)$, and otherwise $R_t$ is irrelevant. Hence, we obtain that $\mathbb E_{{\cal T}_t} [\hat R_t(a)] = \mathbb E_{{\cal T}_t} [\frac{1}{t} \sum_{\tau = 1}^t R_\tau^a] = R(a)$ for all $a$ and $t$.

Note that $\hat R_t(a)$ is a sum of $t$ random variables belonging to the $[0,\frac{1}{\pi_t^{lmin}}]$ interval. By scaling $R(a)$ and $\hat R_t(a)$ by a factor of $\pi_t^{lmin}$ we scale the random variables to the $[0,1]$ interval, where Lemmas \ref{lem:Martin} and \ref{lem:Laplace} can be applied. 

We apply PAC-Bayesian analysis to the scaled version of $R(a)$ and $\hat R_t(a)$ for a fixed $t$:
\begin{align}
t\cdot kl(\pi_t^{lmin} \hat R_t(\rho_t)\|\pi_t^{lmin} R(\rho_t)) &= t\cdot kl(\mathbb E_{\rho_t(a)} [\pi_t^{lmin} \hat R_t(a)] \| \mathbb E_{\rho_t(a)} [\pi_t^{lmin} R(\rho)])\notag\\
&\leq \mathbb E_{\rho_t(a)} [t\cdot kl(\pi_t^{lmin} \hat R_t(a)\| \pi_t^{lmin} R(a))]\label{eq:1}\\
&\leq KL(\rho_t\|\mu_t) + \ln \mathbb E_{\mu_t(a)} [e^{t\cdot kl(\pi_t^{lmin} \hat R_t(a)\|\pi_t^{lmin} R(a))}],\label{eq:2}
\end{align}
where \eqref{eq:1} is due to convexity of $kl$ and \eqref{eq:2} is by Lemma \ref{lem:PAC-Bayes}.

The second term in \eqref{eq:2} can be bounded with high probability:
\begin{align}
\mathbb E_{\mu_t(a)} [e^{t\cdot kl(\pi_t^{lmin} \hat R_t(a)\|\pi_t^{lmin} R(a))}] &\leq \frac{1}{\delta_t} \mathbb E_{{\cal T}_t} \mathbb E_{\mu_t(a)} [e^{t\cdot kl(\pi_t^{lmin} \hat R_t(a)\|\pi_t^{lmin} R(a))}]\label{eq:2-1}\\
&=\frac{1}{\delta_t} \mathbb E_{\mu_t(a)} \mathbb E_{{\cal T}_t} [e^{t\cdot kl(\pi_t^{lmin} \hat R_t(a)\|\pi_t^{lmin} R(a))}]\label{eq:2-2}\\
&\leq \frac{1}{\delta_t} (t+1)\label{eq:2-3},
\end{align}
where \eqref{eq:2-1} holds with probability greater than $1-\delta_t$ by Markov's inequality (Lemma \ref{lem:Markov}), the interchange of expectations in \eqref{eq:2-2} is possible since $\mu_t$ is independent of ${\cal T}_t$, and \eqref{eq:2-3} is by Lemma \ref{lem:Martin} and Lemma \ref{lem:Laplace}. Substitution of \eqref{eq:2-3} into \eqref{eq:2} yields with probability greater than $1-\delta_t$:
\[
kl(\pi_t^{lmin} \hat R_t(\rho_t)\|\pi_t^{lmin} R(\rho_t)) \leq \frac{KL(\rho_t\|\mu_t) + \ln \frac{t+1}{\delta_t}}{t}.
\]

Finally, by setting $\delta_t = \frac{\delta}{t(t+1)} \geq \frac{\delta}{(t+1)^2}$ and applying union bound we obtain \eqref{eq:PAC-Bayes-Bandits} for all $t$ simultaneously (it is well-known that $\sum_{t=1}^\infty \frac{1}{t(t+1)} = \sum_{t=1}^\infty \left (\frac{1}{t} - \frac{1}{t+1}\right) = 1$).
\end{proof}

The key ingredient that made the proof of Theorem \ref{thm:PAC-Bayes-Bandits} possible was Lemma \ref{lem:Martin}, which enabled us to bound $\mathbb E_{{\cal T}_t} [e^{t\cdot kl(\pi_t^{lmin} \hat R_t(a)\|\pi_t^{lmin} R(a))}]$ even though the variables $\{R_1^a,..,R_t^a\}$ are dependent.

\section{Proof of Theorem \ref{thm:PAC-Bayes-Martin} (PAC-Bayesian Analysis Based on Hoeffding-Azuma Inequality)}
\label{sec:Martin}

In this section we provide an alternative PAC-Bayesian bound for $|\hat R_t^{w^t}(\rho_t) - R(\rho_t)|$ by using Hoeffding-Azuma inequality. 

\begin{proof}[Proof of Theorem \ref{thm:PAC-Bayes-Martin}]
Let
\[
M_t^i(a) = \frac{1}{t} \sum_{\tau = 1}^i w_\tau^t (R_\tau^a - R(a)).
\]
Observe that $M_t^1(a),..,M_t^t(a)$ is a martingale [since $\mathbb E_{R_i^a} \left [M_t^i(a)\right ] = M_t^{i-1}(a)$] and \linebreak[4] $M_t^t(a) = \left (\sum_{\tau = 1}^t w_\tau^t \right )(\hat R_t^{w^t}(a) - R(a))$. Note that $(M_t^i - M_t^{i-1}) \in [-\frac{1}{\pi_i^{min}}, \frac{1}{\pi_i^{min}}]$ and $\mathbb E M_t^t = 0$. Hence, by Hoeffding-Azuma inequality (Lemma \ref{lem:Azu}), for all $a$:
\[
\mathbb E_{{\cal T}_t} \left [e^{\lambda_t \left (\sum_{\tau = 1}^t w_\tau^t \right ) (\hat R_t^{w^t}(a) - R(a))} \right ] = \mathbb E_{{\cal T}_t} \left [e^{M_t^t(a)} \right ] \leq e^{\frac{1}{2} \lambda_t^2 \sum_{\tau = 1}^t \left (\frac{w_\tau^t}{\pi_\tau^{min}}\right)^2}.
\]

By going back to the proof of Theorem \ref{thm:PAC-Bayes-Bandits} and replacing $kl(\pi_t^{lmin} \hat R_t(a)\|\pi_t^{lmin} R(a))$ with $\hat R_t^{w^t}(a) - R(a)$ and substituting the bound on $\mathbb E_{{\cal T}_t} [ e^{t \cdot kl(\pi_t^{lmin} \hat R_t(a)\|\pi_t^{lmin} R(a))} ]$ with the bound on \linebreak[4] $\mathbb E_{{\cal T}_t} [ e^{\lambda_t \left (\sum_{\tau = 1}^t w_\tau^t \right ) (\hat R_t^{w^t}(a) - R(a))} ]$ we derived above we obtain that with probability greater than $1-\frac{1}{2}\delta$ for all $\rho_t$
\[
\hat R_t^{w^t}(\rho_t) - R(\rho_t) \leq \frac{KL(\rho_t\|\mu_t) + \frac{1}{2} \lambda_t^2 \sum_{\tau = 1}^t \left (\frac{w_\tau^t}{\pi_\tau^{min}}\right)^2 + 2 \ln(t+1) + \ln \frac{2}{\delta}}{\lambda_t \sum_{\tau = 1}^t w_\tau^t}
\]
and, by a symmetric argument applied to $-M_t^1(a),..,-M_t^t(a)$,
\[
R(\rho_t) - \hat R_t^{w^t}(\rho_t) \leq \frac{KL(\rho_t\|\mu_t) + \frac{1}{2} \lambda_t^2 \sum_{\tau = 1}^t \left (\frac{w_\tau^t}{\pi_\tau^{min}}\right)^2 + 2 \ln(t+1) + \ln \frac{2}{\delta}}{\lambda_t \sum_{\tau = 1}^t w_\tau^t}.
\]
Hence, both hold simultaneously with probability greater than $1-\delta$ and yield \eqref{eq:PAC-Bayes-Martin}.
\end{proof}

\section{Proof of Theorem \ref{thm:regret} (The Regret Bound)}
\label{sec:regret}

In this section we derive a regret bound based on Theorem \ref{thm:PAC-Bayes-Bandits}. We then discuss some possible ways to tighten the regret bound. 

The regret bound is derived for the special kind of posterior distribution $\tilde \rho_t^{_{exp}}$ defined in \eqref{eq:tilderho} in Theorem \ref{thm:regret}, which is used as sampling distribution $\pi_{t+1}$ for the next round of the game, as described in the theorem. Furthermore, we define a special kind of prior distribution $\mu_t^{_{exp}}$ as:
\begin{equation}
\mu_t^{_{exp}}(a) = \frac{1}{Z(\mu_t^{_{exp}})} e^{\gamma_t R(a)}.
\label{eq:mu}
\end{equation}
The prior $\mu_t^{_{exp}}$ depends on the true expected rewards $R(a)$, but not on the sample and hence it is a legal prior.

\begin{proof}[Proof of Theorem \ref{thm:regret}]
Let $a^*$ be the action with the highest reward. The expected regret of the prediction strategy $\tilde \rho_t^{_{exp}}$ at step $t+1$ can be written as follows:
\begin{equation}
R(a^*) - R(\tilde \rho_t^{_{exp}}) = [R(a^*) - \hat R_t(a^*)] + [\hat R_t(a^*) - \hat R_t(\rho_t^{_{exp}})] + [\hat R_t(\rho_t^{_{exp}}) - R(\rho_t^{_{exp}})] + [R(\rho_t^{_{exp}}) - R(\tilde \rho_t^{_{exp}})].\label{eq:regret-decomposition}
\end{equation}
We bound the terms in \eqref{eq:regret-decomposition} one-by-one.

$R(a^*)$ and $\hat R_t(a^*)$ are the expected and the empirical rewards of a prediction strategy, which is a delta distribution on $a^*$. Hence, by Theorem \ref{thm:PAC-Bayes-Bandits}:
\begin{align}
R(a^*) - \hat R_t(a^*) &\leq \frac{1}{\varepsilon_t} \sqrt{\frac{-\ln \mu_t^{_{exp}}(a^*) + 3 \ln(t+1) - \ln \delta}{2t}}\notag\\
&=\frac{1}{\varepsilon_t} \sqrt{\frac{\ln \frac{Z(\mu_t^{_{exp}})}{e^{\gamma_t R(a^*)}} + 3 \ln(t+1) - \ln \delta}{2t}}\notag\\
&\leq \frac{1}{\varepsilon_t} \sqrt{\frac{\ln (K) + 3 \ln(t+1) - \ln \delta}{2t}},\label{eq:maxR}
\end{align}
where in \eqref{eq:maxR} we used the fact that $R(a^*) \geq R(a)$ for all $a$ and hence $e^{\gamma_t R(a^*)} \geq \frac{1}{K} \sum_a e^{\gamma_t R(a)} = \frac{1}{K} Z(\mu_t^{_{exp}})$.

For the second term in \eqref{eq:regret-decomposition} we write:
\begin{align}
\hat R_t(a^*) - \hat R_t(\rho_t^{_{exp}}) &= \sum_a (\hat R_t(a^*) - \hat R_t(a)) \rho_t^{_{exp}}(a)\notag\\
&= \sum_a (\hat R_t(a^*) - \hat R_t(a)) \frac{e^{\gamma_t \hat R_t(a)}}{Z(\rho_t^{_{exp}})}\notag\\
&= \sum_a (\hat R_t(a^*) - \hat R_t(a)) \frac{e^{-\gamma_t (\hat R_t(a^*) - \hat R_t(a))}}{\sum_{a'}e^{-\gamma_t (\hat R_t(a^*) - \hat R_t(a'))}}\notag\\
&\leq \frac{K}{\gamma_t},\label{eq:exp}
\end{align}
where in \eqref{eq:exp} follows from the technical lemma below. The proof of the lemma is provided at the end of this section.
\begin{lemma}
\label{lem:expsum}
Let $x_1 = 0$ and $x_2,..,x_n$ be $n-1$ arbitrary numbers. For any $\alpha > 0$ and $n \geq 2$$:$
\[
\frac{\sum_{i=1}^n x_i e^{-\alpha x_i}}{\sum_{j=1}^n e^{-\alpha x_j}} \leq \frac{n}{\alpha}.
\]
\end{lemma}

The third term in \eqref{eq:regret-decomposition} is bounded by the following lemma adapted from \citet{LLST10}. The proof of this lemma is also provided at the end of this section.
\begin{lemma}
\label{lem:KLKL}
For $\mu_t^{_{exp}}$ and $\rho_t^{_{exp}}$ defined by \eqref{eq:mu} and \eqref{eq:rho} under the conditions of Theorem \ref{thm:PAC-Bayes-Bandits} the following holds simultaneously with the assertion of Theorem \ref{thm:PAC-Bayes-Bandits}$:$
\begin{equation}
\left |\hat R_t(\rho_t^{_{exp}}) - R(\rho_t^{_{exp}}) \right |\leq \frac{1}{\varepsilon_t\sqrt{2 t}} \left (\frac{\gamma_t}{\varepsilon_t\sqrt{2 t}} + \sqrt{3 \ln(t+1) - \ln \delta} \right).
\label{eq:PAC-Bayes-Bandits-L1-KL}
\end{equation}
\end{lemma}

Finally, for the last term in \eqref{eq:regret-decomposition}:
\begin{align}
R(\rho_t^{_{exp}}) - R(\tilde \rho_t^{_{exp}}) &= \sum_a (\rho_t^{_{exp}}(a) - \tilde \rho_t^{_{exp}}(a)) R(a)\notag\\
&\leq \frac{1}{2} \|\rho_t^{_{exp}} - \tilde \rho_t^{_{exp}}\|_1\label{eq:rho-tilde-rho}\\
&=\frac{1}{2} \sum_a \left |\rho_t^{_{exp}}(a) - (1 - K \varepsilon_{t+1}) \rho_t^{_{exp}}(a) - \varepsilon_{t+1} \right |\notag\\
&= \frac{1}{2} \sum_a \left | K \varepsilon_{t+1} \rho_t^{_{exp}}(a) - \varepsilon_{t+1} \right |\notag\\
&\leq \frac{1}{2} K \varepsilon_{t+1} \sum_a \rho_t^{_{exp}}(a) + \frac{1}{2} K \varepsilon_{t+1}\notag\\
&= K \varepsilon_{t+1}.\notag
\end{align}
In \eqref{eq:rho-tilde-rho} we used the fact that $R(a)$ is bounded by 1 and $\rho_t^{_{exp}}$ and $\tilde \rho_t^{_{exp}}$ are probability distributions.

Gathering all the terms and substituting them back into \eqref{eq:regret-decomposition} we obtain:
\begin{align*}
R(a^*) - R(\tilde \rho_t^{_{exp}}) \leq \frac{1}{\varepsilon_t} \sqrt{\frac{\ln (K) + 3 \ln(t+1) - \ln \delta}{2t}} &+ \frac{K}{\gamma_t}\notag\\
& + \frac{1}{\varepsilon_t\sqrt{2 t}} \left (\frac{\gamma_t}{\varepsilon_t\sqrt{2 t}} + \sqrt{3 \ln(t+1) - \ln \delta} \right) + K \varepsilon_{t+1}.
\end{align*}

By choosing $\gamma_t = K^{1/4}t^{1/4}$ and $\varepsilon_t = K^{-1/4}t^{-1/4}$ we get:
\[
R(a^*) - R(\tilde \rho_t^{_{exp}}) \leq \frac{K^{3/4}}{(t+1)^{1/4}}
\left (
\sqrt{\frac{\ln(K) + 3 \ln(t+1) - \ln \delta}{2K}} + 1 + \frac{1}{2} + \sqrt{\frac{3 \ln(t+1) - \ln \delta}{2K}} + 1 
\right ).
\]

By integration over $t$ the total regret is bounded by $\tilde O(K^{3/4}t^{3/4})$, where $\tilde O$ hides logarithmic factors.
\end{proof}

\subsection{Proofs of Technical Lemmas for Section \ref{sec:regret}}
\label{sec:KL}

We conclude this section with proofs of the two technical lemmas used in the proof of the regret bound.

\begin{proof}[Proof of Lemma \ref{lem:expsum}]
Since $x_1 = 0$ we have:
\begin{align*}
\frac{\sum_{i=1}^n x_i e^{-\alpha x_i}}{\sum_{j=1}^n e^{-\alpha x_j}} &= \frac{\sum_{i=1}^n x_i e^{-\alpha x_i}}{1 + \sum_{j=2}^n e^{-\alpha x_j}}\\
&\leq \sum_{i=1}^n x_i e^{-\alpha x_i}\\
&\leq \frac{n}{\alpha},
\end{align*}
where the last inequality follows from the fact that $x e^{-\alpha x} \leq \frac{1}{\alpha}$.
\end{proof}

We note that by numerical simulations it seems that a tighter bound $\frac{\sum_{i=1}^n x_i e^{-\alpha x_i}}{\sum_{j=1}^n e^{-\alpha x_j}} \leq \frac{\ln(K)}{\alpha}$ holds, but we were unable to prove it analytically.

The proof of Lemma \ref{lem:KLKL} is adapted with minor modifications from \citet{LLST10} and is based on the following two lemmas, which are also adapted from \citet{LLST10} and are proved right after the proof of Lemma \ref{lem:KLKL}.

\begin{lemma}
\label{lem:KL}
For $\mu_t^{_{exp}}$ and $\rho_t^{_{exp}}$ defined by \eqref{eq:mu} and \eqref{eq:rho}$:$
\begin{equation}
\label{eq:KL}
KL(\rho_t^{_{exp}}\|\mu_t^{_{exp}}) \leq \gamma_t \left ([\hat R_t(\rho_t^{_{exp}}) - R(\rho_t^{_{exp}})] + [R(\mu_t^{_{exp}}) - \hat R_t(\mu_t^{_{exp}})] \right ).
\end{equation}
\end{lemma}
\begin{lemma}
\label{lem:KLbound}
For $\mu_t^{_{exp}}$ and $\rho_t^{_{exp}}$ defined by \eqref{eq:mu} and \eqref{eq:rho} under the conditions of Theorem \ref{thm:PAC-Bayes-Bandits} the following holds simultaneously with the assertion of Theorem \ref{thm:PAC-Bayes-Bandits}$:$
\begin{equation}
KL(\rho_t^{_{exp}}\|\mu_t^{_{exp}}) \leq \left (\frac{\gamma_t}{\varepsilon_t\sqrt{2 t}}\right )^2 + 2 \left (\frac{\gamma_t}{\varepsilon_t\sqrt{2 t}} \right )\sqrt{3 \ln(t+1) - \ln \delta}.
\label{eq:KLbound}
\end{equation}
\end{lemma}

\begin{proof}[Proof of Lemma \ref{lem:KLKL}] Substitution of \eqref{eq:KLbound} into \eqref{eq:PAC-Bayes-Bandits-L1} yields \eqref{eq:PAC-Bayes-Bandits-L1-KL}.
\end{proof}

\begin{proof}[Proof of Lemma \ref{lem:KL}]
\begin{align}
KL(\rho_t^{_{exp}}\|\mu_t^{_{exp}}) &= \sum_a \rho_t^{_{exp}}(a) \ln \left (\frac{e^{\gamma_t \hat R_t(a)} Z(\mu_t^{_{exp}})}{e^{\gamma_t R(a)} Z(\rho_t^{_{exp}})} \right )\notag\\
&= \sum_a \rho_t^{_{exp}}(a) \gamma_t (\hat R_t(a) - R(a)) - \ln \left (\frac{\sum_a e^{\gamma_t \hat R_t(a)}}{Z(\mu_t^{_{exp}})}\right )\notag\\
&= \gamma_t [\hat R_t(\rho_t^{_{exp}}) - R(\rho_t^{_{exp}})] - \ln \left (\sum_a \mu_t^{_{exp}}(a) e^{\gamma_t (\hat R_t(a) - R(a))}\right )\label{eq:4-1}\\
&\leq \gamma_t \left ([\hat R_t(\rho_t^{_{exp}}) - R(\rho_t^{_{exp}})] + [R(\mu_t^{_{exp}}) - \hat R_t(\mu_t^{_{exp}})] \right ).\label{eq:4-2}
\end{align}
In \eqref{eq:4-1} we used the fact that $\frac{1}{Z(\mu_t^{_{exp}})} = \mu_t^{_{exp}}(a) e^{-\gamma_t R(a)}$ (for any $a$) and in \eqref{eq:4-2} we used the concavity of $\ln$.
\end{proof}

\begin{proof}[Proof of Lemma \ref{lem:KLbound}]
By Theorem \ref{thm:PAC-Bayes-Bandits} and simultaneously with it we have:
\begin{align*}
\hat R_t(\rho_t^{_{exp}}) - R(\rho_t^{_{exp}}) &\leq \frac{1}{\varepsilon_t} \sqrt{\frac{KL(\rho_t^{_{exp}}\|\mu_t^{_{exp}}) + 3 \ln(t+1) - \ln \delta}{2t}}\\
R(\mu_t^{_{exp}}) - \hat R_t(\mu_t^{_{exp}}) &\leq \frac{1}{\varepsilon_t} \sqrt{\frac{3 \ln(t+1) - \ln \delta}{2t}}.
\end{align*}
By substituting this into \eqref{eq:KL} we have:
\[
KL(\rho_t^{_{exp}}\|\mu_t^{_{exp}}) \leq \frac{\gamma_t}{\varepsilon_t \sqrt{2t}} \sqrt{KL(\rho_t^{_{exp}}\|\mu_t^{_{exp}}) + 3 \ln(t+1) - \ln \delta} + \frac{\gamma_t}{\varepsilon_t \sqrt{2t}} \sqrt{3 \ln(t+1) - \ln \delta}.
\]
If $KL(\rho_t^{_{exp}}\|\mu_t^{_{exp}}) \leq \frac{\gamma_t}{\varepsilon_t \sqrt{2t}}$ we are done. Otherwise, by rearranging the terms we obtain:
\begin{align*}
(KL(\rho_t^{_{exp}}\|\mu_t^{_{exp}}))^2 - 2 KL(\rho_t^{_{exp}}\|\mu_t^{_{exp}}) \frac{\gamma_t}{\varepsilon_t \sqrt{2 t}} \sqrt{3 \ln(t+1) - \ln \delta} + \left (\frac{\gamma_t}{\varepsilon_t\sqrt{2t}}\right )^2 (3 \ln(t+1) - \ln \delta)&\\
\leq \left (\frac{\gamma_t}{\varepsilon_t\sqrt{2t}}\right )^2 KL(\rho_t^{_{exp}}\|\mu_t^{_{exp}}) + \left (\frac{\gamma_t}{\varepsilon_t\sqrt{2t}}\right )^2 (3 \ln(t+1) &- \ln \delta),
\end{align*}
which together with the fact that $KL(\rho_t^{_{exp}}\|\mu_t^{_{exp}}) \geq 0$ implies the result.
\end{proof}

\section{Discussion}
\label{sec:disc}

We presented a lemma that allows to bound expectations of convex functions of certain sequentially dependent variables by expectations of the same functions of i.i.d. Bernoulli variables. We showed that this lemma can be used to derive an alternative to Hoeffding-Azuma inequality for convergence of martingale values.

We presented two different approaches to PAC-Bayesian analysis of martingale-type sequentially dependent random variables, which was an important challenge for PAC-Bayesian analysis for a long time. Our contribution opens the possibility to apply PAC-Bayesian analysis in multiple domains, where sequentially dependent variables are encountered. For example, Theorems \ref{thm:PAC-Bayes-Bandits} and \ref{thm:PAC-Bayes-Martin} can be used to bound convergence of uncountable number of parallel martingale sequences, where simple union bound does not apply.

We answered positively an important open question whether PAC-Bayesian analysis can be applied under limited feedback and used to study the exploration-exploitation trade-off. Although our regret bound for the multiarmed bandit problem is far from state-of-the-art yet, we believe that this gap can be closed in future work.

Multiarmed bandits are just the first tier in a whole hierarchy of reinforcement learning problems with increasing structural complexity, including continuum-armed bandits, contextual bandits, and reinforcement learning in discrete and continuous spaces. In many of these domains Bayesian approaches and incorporation of prior knowledge have already proved beneficial in practice, but their rigorous analysis remains difficult to carry out. We believe that PAC-Bayesian approach will prove to be as useful for this purpose as it already proved itself in the domain of supervised learning.

\section*{Acknowledgements}

We thank John Langford for helpful discussions at the early stages of this work and Andreas Maurer for his comments on this manuscript. We are also grateful to anonymous reviewers for their insightful comments and useful references. This work was supported in part by the IST Programme of the European Community, under the PASCAL2 Network of Excellence, IST-2007-216886. This publication only reflects the authors' views.

\bibliography{bibliography}



\end{document}